\documentclass[11pt]{article}

\usepackage[preprint]{acl}

\usepackage{times}
\usepackage{latexsym}

\usepackage[T1]{fontenc}

\usepackage[utf8]{inputenc}

\usepackage{microtype}
\usepackage{amssymb}

\usepackage{inconsolata}

\usepackage{graphicx}
\usepackage{booktabs}
\usepackage[table]{xcolor}

\title{Where Does the Answer Come From? Benchmarking View-Level Visual Evidence Identification in Multi-View MLLMs for Autonomous Driving}

\author{
{\bf Yimu Wang}\thanks{Equal contribution.}, 
{\bf Yee Man Choi}\footnotemark[1], 
{\bf Barry Zhang, Mozhgan Nasr Azadani,}\\
{\bf Sean Sedwards, and Krzysztof Czarnecki}\\
University of Waterloo
}

\usepackage[utf8]{inputenc}
\usepackage[T1]{fontenc}
\usepackage{booktabs}
\usepackage{xcolor}
\usepackage[most]{tcolorbox}

\definecolor{bordergray}{RGB}{200, 200, 200}
\definecolor{bglight}{RGB}{247, 247, 247}
\definecolor{titlebg}{RGB}{230, 230, 230}

\newtcolorbox{TemplateBox}[1]{
    enhanced,
    colback=bglight,
    colframe=bordergray,
    boxrule=1.5pt,
    arc=10pt,
    boxsep=10pt,
    top=15pt,
    fontupper=\ttfamily\small,
    fonttitle=\bfseries,
    coltitle=black,
    attach boxed title to top left={
        xshift=20pt,
        yshift=-10pt
    },
    boxed title style={
        enhanced,
        colback=titlebg,
        colframe=bordergray,
        boxrule=1.5pt,
        arc=6pt
    },
    title={#1}
}

\usepackage{cleveref}

\begin{document}
\maketitle
\begin{abstract}
Multimodal large language models (MLLMs) achieve strong results on visual reasoning benchmarks, but answer accuracy alone does not indicate whether a model relied on the correct visual evidence.
This gap is particularly important in multi-view driving scenes used for autonomous driving, where a model can produce a plausible answer while grounding it in the wrong camera view.
We introduce a multi-view visual question answering benchmark for evaluating evidence-source identification: given six synchronized NuScenes views and a question, the model must identify the supporting camera view and answer the question.
The benchmark contains 122 conflict-centric question-answer pairs from 73 scenes, spanning causality, counterfactual reasoning, and intent prediction.
View labels are proposed by an automatic conflict-mining pipeline and manually verified by annotators.
We evaluate three settings: camera-view selection, oracle QA given the golden view, and joint prediction in which the model selects a view and answers in one pass.
Answers are evaluated in both multiple-choice and free-form formats, using exact match for structured predictions and an LLM judge for free-form responses.
By explicitly separating visual-source identification from answer correctness, the benchmark exposes grounding failures that answer-only evaluation misses.
\end{abstract}

\begin{figure}[t]
\centering
\IfFileExists{pics/teaser.pdf}{
    \includegraphics[width=\linewidth]{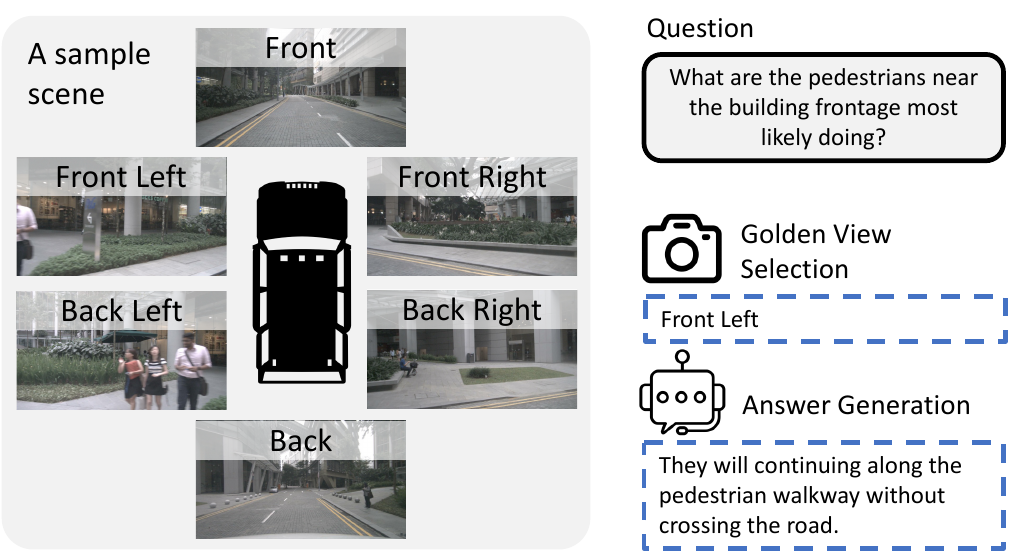}
}{
    \fbox{
    \begin{minipage}[c][0.32\textheight][c]{0.94\linewidth}
    \centering
    \vspace{0.5em}
    \textbf{Overview figure missing}\\[0.5em]
    Six synchronized surround-view cameras around the ego vehicle.\\
    One view is highlighted as the golden view.\\
    The right panel illustrates three evaluations: view selection, oracle QA with the golden view, and joint prediction.\\
    \vspace{0.5em}
    \end{minipage}
    }
}
\vspace{-2em}
\caption{Benchmark overview. Given six synchronized NuScenes camera views and a question, the benchmark evaluates whether an MLLM can identify the supporting camera view and answer the question. The figure illustrates surround-view input and golden-view grounding.
}
\vspace{-1.5em}
\label{fig:overview}
\end{figure}

\section{Introduction}

Multimodal large language models (MLLMs)~\citep{liu_VisualInstructionTuning_2023,liu_ImprovedBaselinesVisual_2023} have shown strong performance on image understanding~\cite{qian_NuScenesQAMultiModalVisual_2024} and visual question answering~\cite{liu_MMBenchYourMultimodal_2025}.
Yet most evaluations emphasize final-answer correctness and pay limited attention to whether the answer is grounded in the right visual source.
Even in single-image settings, models can produce plausible responses~\cite{li_EvaluatingObjectHallucination_2023} by exploiting language priors or dataset shortcuts. 
In multi-view settings~\cite{caesar_NuScenesMultimodalDataset_2020}, this concern becomes more explicit: when given several synchronized views of the same scene, a reliable model should identify which view contains the evidence needed to answer the question.

We study this problem in autonomous driving scenes~\cite{caesar_NuScenesMultimodalDataset_2020,Sun_2020_CVPR,qian_NuScenesQAMultiModalVisual_2024}, where vehicles, pedestrians, and cyclists may appear in different camera views around the ego vehicle. 
Consider a question about a potential conflict with a vehicle approaching from the back-left. 
If a model answers correctly but attributes its evidence to the front camera, it has not demonstrated faithful multi-view grounding. 
Such behaviour may reflect guessing, language bias, or hallucinated evidence attribution, echoing broader concerns about object hallucination and visual illusion in MLLMs~\citep{li_EvaluatingObjectHallucination_2023,guan_HallusionBenchAdvancedDiagnostic_2024}.
This matters in safety-critical driving settings~\cite{zhou_VisionLanguageModels_2024,cui_SurveyMultimodalLarge_2023}, where trustworthy reasoning requires not only a correct answer but also a correct evidence source.

We propose a benchmark for evaluating view identification in multi-view MLLMs, as illustrated in \Cref{fig:overview}.
Each instance contains six synchronized NuScenes camera views~\citep{caesar_NuScenesMultimodalDataset_2020} and a natural-language question.
The central annotation is a manually verified golden view: the camera channel that provides the most direct visual evidence for answering the question. 
We evaluate three complementary settings.
First, in \emph{view selection}, the model receives all six views and predicts only the golden view. 
Second, in \emph{oracle QA}, the model receives the golden-view image and answers the question, isolating reasoning when the correct visual source is given. 
Third, in \emph{joint prediction}, the model receives all six views and predicts both the supporting view and the answer. 
This setup distinguishes genuinely grounded behaviour from cases where models answer correctly while selecting the wrong visual source.

Our contributions are:
\begin{itemize}
    \setlength{\itemsep}{1pt}
    \setlength{\topsep}{2pt}
    \setlength{\parsep}{0pt}
    \setlength{\partopsep}{0pt}
    \item We introduce a compact diagnostic multi-view VQA benchmark for evaluating whether MLLMs can identify the visual evidence source behind their answers.
    \item We construct 122 conflict-centric QA pairs from 73 scenes, covering causality, counterfactual reasoning, intent prediction.
    \item We propose a three-part evaluation protocol covering view selection, oracle QA given the golden view, and joint prediction.
    \item We evaluate both multiple-choice and free-form answer formats, using exact match for structured outputs and an LLM judge for free-form responses.
    \item We benchmark representative proprietary and open-source MLLMs, including GPT, Gemini, Claude, Qwen-VL, and InternVL families, under a unified zero-shot prompting protocol.
\end{itemize}

\section{Related Work}

Existing MLLM benchmarks primarily score final answers \citep{liu_MMBenchYourMultimodal_2025,gurari_VizWizGrandChallenge_2018}, while giving limited insight into whether models use the correct visual evidence.
This issue is especially important in autonomous driving~\cite{cui_SurveyMultimodalLarge_2023}, where information is distributed across synchronized surround-view cameras.
Our benchmark is complementary to answer-centric driving QA~\cite{caesar_NuScenesMultimodalDataset_2020}, including graph-structured driving VQA~\citep{sima_DriveLMDrivingGraph_2024}: we explicitly evaluate whether models can first identify the supporting camera view, then produce a correct answer.
The closest driving VQA settings~\citep{sima_DriveLMDrivingGraph_2024,qian_NuScenesQAMultiModalVisual_2024} usually treat multi-view images as a shared visual context and score the final response, whereas our benchmark makes the evidence source itself an explicit prediction target. 
This distinction is also different from generic multi-image reasoning~\citep{jiang2024mantis}: the candidate evidence sources are calibrated camera channels from the same timestamp, so a wrong source can be objectively checked against a manually verified golden view. 
By evaluating oracle QA and joint prediction separately, we can distinguish failures of visual-source localization from failures of downstream answer reasoning. 
In addition, unlike answer-only benchmarks, our evaluation explicitly identifies wrong-view/correct-answer behaviour, which is central for diagnosing shortcut reasoning and evidence misattribution. 
This shifts evaluation from answer plausibility to evidence-grounded reasoning.

\section{Task, evaluation metric, and dataset}

\subsection{Problem Setup}

\begin{figure*}[t]
\centering
\IfFileExists{pics/dataset_flow.pdf}{
    \includegraphics[width=\linewidth]{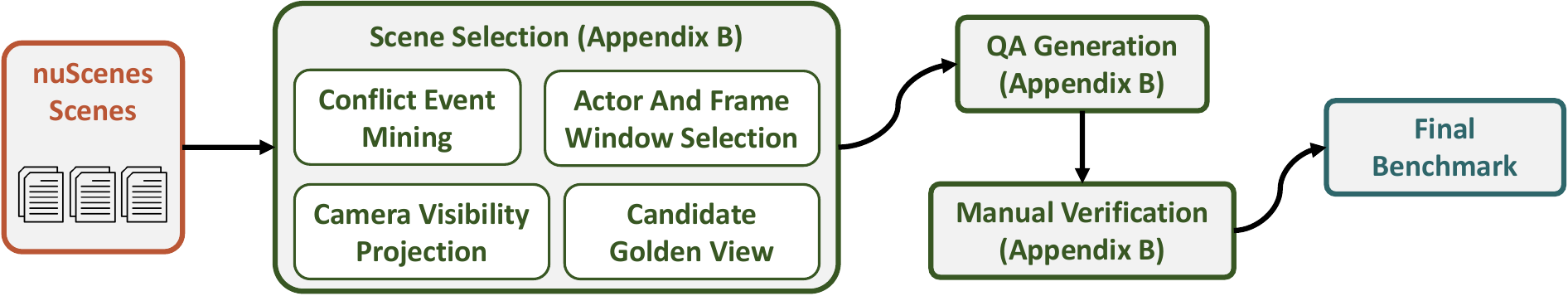}
}{
    \fbox{
    \begin{minipage}[c][0.22\textheight][c]{0.94\linewidth}
    \centering
    \vspace{0.5em}
    \textbf{Dataset generation placeholder}\\[0.5em]
    NuScenes scenes $\rightarrow$ conflict event mining $\rightarrow$ actor and frame-window selection\\
    $\rightarrow$ camera visibility projection $\rightarrow$ candidate golden view\\
    $\rightarrow$ QA generation $\rightarrow$ manual verification $\rightarrow$ final benchmark
    \vspace{0.5em}
    \end{minipage}
    }
}
\vspace{-2em}
\caption{Dataset construction pipeline. We mine conflict-centric candidate events from NuScenes, identify involved actors and event frame windows, project actor 3D boxes into the six camera views to propose candidate golden views, generate QA pairs, and manually verify answerability, single-view grounding, and reference answers.}
\label{fig:dataset-pipeline}
\vspace{-1em}
\end{figure*}

Each example contains six synchronized NuScenes camera views~\citep{caesar_NuScenesMultimodalDataset_2020}, a question $q$, a view label $v^*$, and a reference answer $a^*$.
For answerable examples, $v^*$ is the golden camera view; for unsupported examples, $v^*$ is \textsc{None}.
We evaluate three settings: (1) \emph{view selection} (predict $\hat{v}$ from all six views), (2) \emph{oracle QA} (predict $\hat{a}$ given only the golden-view image for answerable examples), and (3) \emph{joint prediction} (predict both $\hat{v}$ and $\hat{a}$ from all six views).
In joint prediction, the model must choose the view or abstain before answering, reducing post-hoc evidence attribution.

\subsection{Metrics}

View prediction is scored by exact match against $v^*$:
\begin{equation}
    \mathrm{Acc}_\mathit{view} = \frac{1}{N}\sum_{i=1}^{N}\mathbf{1}[\hat{v}_i = v_i^*].
\end{equation}
For answers, multiple-choice outputs use exact match, while free-form outputs are graded by an LLM judge with a semantic-equivalence rubric, following recent use of LLM-based evaluators for open-ended generation~\citep{liu-etal-2023-g,zheng2023judging}.
For joint prediction, we additionally report strict joint success, where both the view and answer must be correct:
\begin{equation}
    \mathrm{Acc}_\mathit{joint} = \frac{1}{N}\sum_{i=1}^{N}\mathbf{1}[\hat{v}_i = v_i^* \land \hat{a}_i = a_i^*].
\end{equation}
For free-form joint outputs, $\hat{a}_i = a_i^*$ denotes judge-evaluated semantic correctness rather than string identity.

\subsection{Dataset}

The dataset contains 122 conflict-centric QA pairs from 73 NuScenes scenes and uses camera inputs only (no LiDAR/radar/BEV). 
The statistics are summarized in \Cref{tab:dataset-stats}, and the construction pipeline is shown in \Cref{fig:dataset-pipeline}.
All examples are manually filtered to have either one dominant supporting camera view or no camera view that provides sufficient evidence.

\begin{table}[t]
\centering
\small
\begin{tabular}{@{}l r@{}}
\toprule
\textbf{Statistic} & \textbf{Value} \\
\midrule
Dataset source & NuScenes~\citep{caesar_NuScenesMultimodalDataset_2020} \\
Camera views & 6 \\
Scenes & 73 \\
QA pairs & 122 \\
\rowcolor{gray!10} \multicolumn{2}{c}{\textbf{Question Types}} \\
Causality & 52 \\
Counterfactual & 35 \\
Intent Prediction & 35 \\
\bottomrule
\end{tabular}
\vspace{-1em}
\caption{Benchmark statistics.}
\label{tab:dataset-stats}
\vspace{-1em}
\end{table}

Scene selection follows an event-centric pipeline: we first mine candidate conflict events from cached kinematic, geometric, and map features, then identify the camera view(s) in which the relevant actor is visible during the event window.
Candidates are post-processed to keep cases where one view provides the clearest support for the question.
We cover six event families (pedestrian crossing, braking, cut-in, lane change, turning, and Car-to-Car Front Turn-Across-Path) to encourage diverse interaction patterns rather than single-object recognition. 
Details are deferred to \Cref{app:scene-selection}.

Question-answer generation targets three reasoning types: causality, counterfactual reasoning, and intent prediction.
Each question can be instantiated in multiple-choice and free-form variants.
This paired design enables deterministic answer scoring for multiple-choice while still testing unconstrained generation quality in free-form evaluation.

Golden views are proposed automatically and manually verified.
Annotators inspect all six synchronized views to confirm that (i) the question is answerable, (ii) the proposed golden view contains the critical evidence, and (iii) no alternative view offers equally direct evidence.
Ambiguous or weakly grounded samples are revised or removed.

\paragraph{Annotation protocol and reliability.}
When multiple views partially show an actor, annotation guidelines prioritize the view with the most direct evidence needed to answer the question; if no single view is dominant, the sample is excluded. 
We also preserve 5 QA pairs whose questions cannot be supported by the camera views to test rejection ability. 
The details of the annotation protocol are provided in Appendix~\ref{app:details}.

To reduce annotation leakage into evaluation, we treat the dataset as test-only and do not create a training split.
This choice matches the benchmark's diagnostic goal: we want to measure whether existing MLLMs can identify the evidence source under realistic multi-view prompting, rather than optimize a model specifically for this label space.
Because the benchmark is compact, the reported numbers should be interpreted as controlled diagnostics of grounding behaviour rather than leaderboard-style saturated performance.
Extended algorithmic details and implementation specifics are provided in Appendix~\ref{app:details}.

\section{Experiment}

\subsection{Models}

We evaluate representative proprietary and open-source MLLMs.
The proprietary group includes GPT-5.4, Gemini, and Claude.
The open-source group includes Qwen2.5-VL~\citep{bai_QwenVLVersatileVisionLanguage_2023,bai_Qwen25VLTechnicalReport_2025}, Qwen3-VL~\citep{bai_Qwen3VLTechnicalReport_2025}, and InternVL3~\cite{zhu_InternVL3ExploringAdvanced_2025}.
All models are tested in a zero-shot setting without task-specific fine-tuning.

\subsection{Experimental Setup}

All models are evaluated under a unified zero-shot protocol with fixed prompts per setting and deterministic decoding where applicable (details in Appendix~\ref{app:experimental-protocol}).
We use three settings: \textbf{view selection} (six synchronized views $\rightarrow$ supporting camera channel), \textbf{oracle QA} (golden-view image only $\rightarrow$ answer, in multiple-choice or free-form format), and \textbf{joint prediction} (six views $\rightarrow$ view, answer, and rationale in one pass, in both answer formats).
Multiple-choice answers (MC) are scored by exact identifier match; free-form answers use a fixed LLM judge for semantic correctness~\cite{liu-etal-2023-g,zheng2023judging}.
In the joint setting, we report both answer quality and strict joint correctness (view and answer both correct).
For unsupported examples, \textsc{None} is the correct view-selection target and oracle QA is not applicable.
Full prompt templates are in \Cref{app:prompts}.

\subsection{Main Results}

\begin{table}[t]
\centering
\small
\begin{tabular}{@{}l c c c@{}}
\toprule
\textbf{Model} & \textbf{Mean (\%)} & \textbf{SD (\%)} & \textbf{95\% CI (\%)}\\
\midrule
GPT-5.4 & 77.54 & 0.45 & [76.98, 78.10] \\
Gemini & 74.10 & 1.89 & [71.76, 76.44]  \\
Claude & 82.62 & 0.69 & [81.77, 83.47]  \\
\midrule
Qwen2.5VL-7B & 12.62 & 1.24 & [11.08, 14.17] \\
Qwen3VL-8B & 61.64 & 2.42 & [58.64, 64.64] \\
InternVL3 & 61.48 & 2.09 & [58.88, 64.07] \\
\bottomrule
\end{tabular}
\vspace{-1em}
\caption{View-selection exact-match accuracy summary over 5 runs. `SD' stands for standard deviation and `95\% CI' stands for 95\% confidence interval.}
\label{tab:exact-match-summary}

\centering
\resizebox{\columnwidth}{!}{
\begin{tabular}{l c c c c c}
\toprule
\textbf{Model} & \textbf{View} & \textbf{Oracle} & \textbf{Oracle} & \textbf{Joint} & \textbf{Joint} \\
 & \textbf{Acc.} & \textbf{MC} & \textbf{Free} & \textbf{MC} & \textbf{Free} \\
\midrule
GPT-5.4 & 77.5 & 86.9 & 79.5 & 66.9 & 61.3 \\
Gemini & 74.1 & 89.3 & 73.8 & 68.8 & 56.8 \\
Claude & 82.6 & 89.3 & 73.0 & 73.2 & 59.8 \\
\midrule
Qwen2.5VL-7B & 12.6 & 78.7 & 37.7 & 8.4 & 4.0 \\
Qwen3VL-8B & 61.6 & 66.4 & 51.4 & 43.5 & 29.0 \\
InternVL3 & 61.5 & 82.0 & 40.2 & 51.7 & 25.4 \\
\bottomrule
\end{tabular}
}
\vspace{-1em}
\caption{Main evaluation summary. View Acc. is exact-match accuracy for view selection. Oracle MC and Oracle Free evaluate answering when the golden view is provided. Joint MC and Joint Free report strict joint correctness, requiring both the selected view and answer to be correct.}
\label{tab:main-results}
\vspace{-1em}
\end{table}

\Cref{tab:exact-match-summary} shows that view selection is stable across repeated runs for the strongest proprietary models, with Claude achieving the highest mean accuracy (82.62\%, 95\% CI: [81.77, 83.47]) and GPT-5.4/Gemini following at 77.54\%/74.10\%.
Open-source performance is generally lower: Qwen3VL-8B and InternVL3 cluster around 61.5\%, while Qwen2.5VL-7B drops to 12.62\%, near or below a naive uniform baseline over the candidate labels.
This gap suggests that multi-view evidence localization remains difficult even when the candidate camera set is small.

\Cref{tab:main-results} further reveals a consistent oracle-to-joint drop across all models (e.g., Claude: 89.3$\rightarrow$73.2 in MC; GPT-5.4: 86.9$\rightarrow$66.9).
Because the oracle setting supplies the golden view, this drop indicates that view identification is a major bottleneck for end-to-end grounded answering.
Finally, free-form scores are consistently below multiple-choice in both oracle and joint settings, suggesting that unconstrained generation introduces additional reasoning, calibration, and grounding errors beyond option selection.

\subsection{Analysis}

We analyze several representative failure modes.
First, models may output a correct answer while selecting a wrong view, indicating shortcut reasoning or evidence misattribution that answer-only metrics would miss.
Second, models may exhibit front-camera bias, over-selecting \texttt{CAM\_FRONT} even when key evidence appears in side or rear channels.
Third, models may confuse adjacent views (e.g., \texttt{CAM\_FRONT} versus \texttt{CAM\_FRONT\_LEFT}) when evidence spans boundary regions.
Fourth, models may fail to abstain on unsupported examples, selecting a visually plausible but non-evidential camera view.
Finally, some free-form answers remain plausible but weakly grounded in the selected image, especially in counterfactual and intent questions.
We analyze per-region confusion patterns, and per-question-type breakdowns, together with qualitative case studies (Appendix~\ref{app:qualitative-cases}).

\section{Conclusion}

We have introduced a multi-view VQA benchmark for autonomous driving that tests whether MLLMs can identify the camera view that supports their answers.
The benchmark evaluates view selection, oracle QA given the golden view, and joint prediction with both multiple-choice and free-form answer formats.
By separating evidence-source identification from answer quality, the benchmark reveals grounding failures hidden by answer-only evaluation.
This enables more targeted analysis of hallucination, evidence attribution, and multi-view reasoning in autonomous driving scenes.

\section{Limitations}

The current benchmark is compact (122 QA pairs) and intended as a diagnostic test rather than a large-scale training resource.
Because all examples are conflict-centric, it does not measure general driving VQA performance.
The distribution of event types and golden views may reflect biases in the conflict-mining pipeline and in NuScenes itself.
Each question currently has a single golden view, although real scenarios may require multi-view evidence.
Free-form answer quality depends on an LLM judge and may therefore include evaluation noise.
Finally, the benchmark currently uses camera images only and excludes LiDAR, radar, and BEV inputs.
Accordingly, our claims focus on view-level evidence attribution in selected conflict scenarios, not on end-to-end autonomous driving competence. 
Because our benchmark is based on NuScenes, both open-source and proprietary MLLMs may have encountered related images, metadata, or derived captions during pretraining, which could bias absolute performance estimates.
For this reason, the benchmark is best interpreted as a controlled diagnostic probe rather than evidence of deployment readiness.

\section{Ethical Considerations}

The benchmark is derived from the public NuScenes dataset and should be used under the original dataset license and terms of use.
Although the images are public research data, they depict real traffic scenes and may include pedestrians, cyclists, and vehicles.
We therefore recommend using the benchmark only for research on model evaluation and not for identifying individuals, inferring private attributes, or supporting deployed driving decisions.

\bibliography{custom}

\clearpage
\newpage
\appendix

\section*{Appendix Overview}

The appendix provides additional context and implementation details for the benchmark.
\Cref{app:extended-related-work} expands the related-work discussion.
\Cref{app:details} provides extended task definitions, dataset construction details, evaluation protocol, annotation procedure, and analysis tables.
\Cref{app:scene-selection} describes the scene-selection algorithm used to mine conflict-centric NuScenes events.
\Cref{app:qualitative-cases} presents qualitative case studies referenced in the main paper.
\Cref{app:prompts} documents the prompts used for view selection, oracle QA, joint prediction, and free-form judging.
\Cref{app:ai-disclosure} provides the AI disclosure statement.

\section{Extended Related Work}
\label{app:extended-related-work}

This section expands the related-work context from the main paper.

\paragraph{Evidence-grounded evaluation for MLLMs.}
A common limitation of multimodal benchmarks~\cite{liu_MMBenchYourMultimodal_2025} is that they evaluate the final answer but not whether the model used the correct visual source.
Recent work on object hallucination and visual illusion has shown that fluent responses can be only weakly tied to the input evidence~\citep{li_EvaluatingObjectHallucination_2023,guan_HallusionBenchAdvancedDiagnostic_2024}.
Our benchmark follows this line of motivation but operationalizes it at the camera-view level: the model must identify which synchronized view supports its answer.

\paragraph{Driving-oriented vision-language benchmarks.}
Driving VLM benchmarks~\cite{qian_NuScenesQAMultiModalVisual_2024,marcu_LingoQAVisualQuestion_2024,li_NuGroundingMultiView3D_2025} typically focus on scene understanding, behavior prediction, risk analysis, or planning-oriented question answering.
Most such settings evaluate answer quality under a fixed input view or under aggregated multi-sensor context, as in graph-structured driving VQA built on driving datasets such as nuScenes \citep{caesar_NuScenesMultimodalDataset_2020,sima_DriveLMDrivingGraph_2024}.
Our setting is complementary because it explicitly evaluates source identification in a surround-view camera setup, where the evidence may appear only in a specific channel.

\paragraph{Multi-image and multi-view reasoning.}
General multi-image reasoning benchmarks~\cite{zhao2024mirb,meng2025mmiu} study cross-image correspondence, temporal consistency, or image-set question answering.
However, many tasks do not require committing to a single evidence source~\citep{jiang2024mantis}.
In contrast, our benchmark is designed around single-dominant-view supervision, making it suitable for diagnosing whether a model can localize evidence before answering.

\paragraph{LLM-as-a-judge for free-form evaluation.}
Open-ended answer evaluation has increasingly relied on model-based judges to score semantic correctness beyond exact string matching.
This approach improves coverage for paraphrases and semantically equivalent responses, but also introduces evaluator sensitivity~\citep{liu-etal-2023-g,zheng2023judging}.
In our setup, judge-based scoring is used only for free-form answers, while multiple-choice outputs remain exact-match, so we can separate answer-format effects from view-selection behavior.

\section{Extended Task, Dataset, and Analysis Details}
\label{app:details}

\subsection{Formal Task Definitions}

Let the six synchronized camera images at one timestamp be $\{I_1,\ldots,I_6\}$, the question be $q$, the view label be $v^*$, and the reference answer be $a^*$.
For answerable examples, $v^*$ is a camera channel; for unsupported examples, $v^*=\textsc{None}$.
We evaluate three settings:
\begin{align}
    f_\mathit{view}(I_1,\ldots,I_6,q) &\rightarrow \hat{v}, \\
    f_\mathit{oracle}(I_{v^*},q) &\rightarrow \hat{a}, \\
    f_\mathit{joint}(I_1,\ldots,I_6,q) &\rightarrow (\hat{v},\hat{a}).
\end{align}
The model may also return a short rationale and visible-evidence text, but the primary evidence-source prediction is the selected view label or \textsc{None}.

\subsection{Experimental Protocol}
\label{app:experimental-protocol}

\paragraph{Prompting and input format.}
Our experimental design has three settings:
\begin{enumerate}
    \item \textbf{Golden (view selection):} the model receives all synchronized views and predicts the supporting camera channel (\texttt{Golden\_view}).
    \item \textbf{Oracle QA:} for answerable examples, the model receives only the golden-view image and answers the question. We evaluate both \textbf{multiple-choice} and \textbf{free-form} variants.
    \item \textbf{Full Loop (joint prediction):} the model receives all six views, selects the supporting view, and answers in one pass (JSON with \texttt{Golden\_view}, \texttt{Answer}, and \texttt{Rationale}). We evaluate both \textbf{multiple-choice} and \textbf{free-form} answer variants.
\end{enumerate}
All prompts explicitly list the candidate camera channels.
The prompt templates also include \textsc{None} as a fallback option when no provided view supports the question.
For unsupported examples, \textsc{None} is scored as the correct view label; for answerable examples, selecting \textsc{None} is scored as incorrect.
If a model outputs an invalid camera name, we map it to the closest valid label only when intent is unambiguous; otherwise, the view prediction is marked incorrect.
Full prompt templates and response schemas are provided in Appendix~\ref{app:prompts}.

\paragraph{Evaluation protocol.}
The overall protocol is summarized in \Cref{tab:prompt-settings}.
For free-form answering, we keep prompting and judge criteria fixed across models so that comparisons primarily reflect model behaviour rather than evaluator drift.
For models with stochastic generation effects, we report repeated-run statistics when available (mean, standard deviation, and 95\% confidence interval).
In the joint setting, strict joint correctness isolates cases where textual correctness is achieved with incorrect visual grounding.

\subsection{Evaluation Details}

View selection is scored by exact match:
\begin{equation}
    \mathrm{Acc}_\mathit{view}=\frac{1}{N}\sum_{i=1}^{N}\mathbf{1}[\hat{v}_i=v_i^*].
\end{equation}
For multiple-choice answers:
\begin{equation}
    \mathrm{Acc}_\mathit{ans}=\frac{1}{N}\sum_{i=1}^{N}\mathbf{1}[\hat{a}_i=a_i^*].
\end{equation}
For joint prediction, as summarized in \Cref{tab:joint-eval}, we also report strict success:
\begin{equation}
    \mathrm{Acc}_\mathit{joint}=\frac{1}{N}\sum_{i=1}^{N}\mathbf{1}[\hat{v}_i=v_i^* \land \hat{a}_i=a_i^*],
\end{equation}
using exact answer match for multiple-choice outputs and judge correctness for free-form outputs.
Oracle QA is computed only on answerable examples because unsupported examples do not have a golden-view image to provide.

\begin{table}[t]
\centering
\small
\resizebox{\columnwidth}{!}{
\begin{tabular}{@{}p{2.35cm}p{2.1cm}p{2.0cm}p{1.55cm}@{}}
\toprule
\textbf{Setting} & \textbf{Visual input} & \textbf{Model output} & \textbf{Metric} \\
\midrule
View selection & 6 views & $c^\star \in \mathcal{C}$ & Exact match \\
Oracle QA & Gold view, answerable subset & $a \in \mathcal{O}$ or text & Exact / judge \\
Joint prediction & 6 views & $(c^\star, a)$ & Exact / judge \\
\bottomrule
\end{tabular}
}
\caption{Inference settings evaluated in our benchmark.
\textit{Oracle QA} supplies the human-annotated camera; \textit{Joint prediction} asks the model to select a view and answer in one pass.}
\label{tab:prompt-settings}
\end{table}

\begin{table}[t]
\centering
\small
\begin{tabular}{l l l}
\toprule
\textbf{View} & \textbf{Answer} & \textbf{Interpretation} \\
\midrule
Correct & Correct & grounded success \\
Correct & Wrong & right source, failed reasoning \\
Wrong & Correct & shortcut or ungrounded answer \\
Wrong & Wrong & full failure \\
\bottomrule
\end{tabular}
\caption{Joint interpretation of view identification and answer quality.}
\label{tab:joint-eval}
\end{table}

\subsection{Benchmark Construction}
\subsubsection{Scene Selection}
\label{app:scene-selection}

The role of scene selection in this pipeline is to mine candidate scenes in which another road user may conflict with, influence, or be influenced by the ego vehicle. For each such scene, our algorithm also identifies the camera view(s) the event was visible in. We then filter these candidate scenes for cases in which the question can be answered from a single dominant camera view. This design allows the benchmark to evaluate view identification with a clean single-label target.

To mine suitable scenes from the dataset, we need something concrete to search for. We therefore define a fixed set of \emph{event categories}, each one a recurring, well-defined class of ego/road user interaction:
\begin{itemize}
    \item \textbf{Pedestrian crossing events}: a pedestrian walks across the trajectory of a vehicle (ego or non-ego).
    \item \textbf{Braking events}: a vehicle exhibits a sustained, significant deceleration.
    \item \textbf{Cut-in events}: a non-ego vehicle moves laterally from an adjacent lane into the ego's lane while remaining ahead of the ego.
    \item \textbf{Lane change events}: a vehicle (ego or non-ego) transitions from one lane to an adjacent lane in the same direction of travel.
    \item \textbf{Left/right turn events}: a vehicle executes a left or right turn at a road intersection.
    \item \textbf{CCFtap events} (Car-to-Car Front Turn-Across-Path): an event between two vehicles at opposite sides of an intersection; one vehicle makes a turn that intersects the path of the oncoming vehicle going straight.
\end{itemize}

The scene-selection algorithm is designed to find events from these categories by querying the nuScenes dataset. The backbone of the algorithm is a pre-populated SQLite database where we cache the per-frame kinematic and geometric features needed to query for events. For a given event category, the algorithm executes a SQL query against this database to select candidates that satisfy conditions that are necessary, but not sufficient, for the event category. The algorithm then runs a post-processing pass to filter candidate events down to final matches. Each final match consists of the involved road users and the frame window during which the event occurs.

For example, when querying for pedestrian crossings, the SQL query retrieves all pedestrian-vehicle pairs whose trajectories cross during a scene, along with the frame each actor reached the crossing point. Post-processing keeps only pairs where these two frames are within a few frames of each other (a near-simultaneous crossing, not minutes apart), and reports the event's frame window as the range between them. A final match might be: in scene-0103, pedestrian P crosses in front of the ego from frame 12 to frame 23.

Because we pre-populate the database once up front, each query just filters cached rows, rather than rescanning the raw nuScenes data each time.

\paragraph{Identifying the relevant camera view.}
To facilitate selecting the golden view for an event, we pre-compute, for each non-ego actor, which of the six ego-mounted cameras it appeared in at each frame.

We pre-compute this camera appearance information by projecting each non-ego actor's 3D bounding box onto each of the six camera image planes. The actor is considered visible in a given camera at a given frame if its projected box overlaps that camera's image rectangle. This per-frame, per-camera visibility information is cached in the SQLite database alongside the other features. The candidate golden view for an event is then simply the camera in which the relevant non-ego actor appeared during the event's frames --- or the sequence of cameras, if the road users were visible in multiple camera views during the scene. This automatic proposal is then filtered manually to keep only cases answerable from a single dominant camera, yielding the benchmark's final golden views.

\paragraph{Per-category criteria.}
The features cached in the SQLite database, all derived from the nuScenes devkit, fall into four groups: (a) per-frame poses for the ego vehicle and every non-ego instance (vehicles, pedestrians), consisting of $(x, y, z)$ position and $(q_w, q_x, q_y, q_z)$ orientation quaternion; (b) 3D bounding boxes for non-ego instances and their per-camera 2D projections; (c) per-frame visibility annotations; and (d) HD-map geometry --- specifically lane connectors (polylines inside an intersection that connect one incoming lane to one outgoing lane) and road intersection polygons.

Below, we describe how each event category is queried:
\begin{itemize}
    \item \textbf{Pedestrian crossing events}: from per-frame poses we determine, for every (vehicle, pedestrian) pair in a scene, whether the pedestrian's trajectory intersects the vehicle's trajectory; in the special case of a stationary or stopped vehicle, we additionally check whether the pedestrian's path passes within a small distance of the vehicle. Pairs satisfying either condition constitute crossings.
    \item \textbf{Braking events}: per-frame speed and acceleration are computed directly from the cached ego and non-ego poses, and a braking event is defined as any window in which a vehicle's deceleration exceeds a sustained magnitude threshold.
    \item \textbf{Cut-in events}: from the cached poses we compute, for each non-ego vehicle, its longitudinal and lateral offsets relative to the ego in the ego's heading frame. A cut-in is defined over a sliding window in which the absolute lateral offset starts sufficiently large (vehicle clearly outside the ego's lane), decreases nearly-monotonically to a sufficiently small value (vehicle clearly inside the ego's lane), and the longitudinal offset remains positive and bounded throughout (the vehicle stays ahead of the ego). 
    \item \textbf{Lane change events}: using lane connectivity derived from the HD map together with the cached vehicle poses, a lane change is defined as a frame window in which a vehicle's pose transitions from one lane polygon to an adjacent, parallel lane polygon.
    \item \textbf{Left/right turn events}: a turn is a special case of an \emph{intersection traversal}, defined as any window during which a vehicle's pose lies inside a road-intersection polygon from the HD map. We classify each traversal as \texttt{left}, \texttt{right}, or \texttt{straight} using two signals: (i) the vehicle's yaw rate across the traversal window, and (ii) the shape of the route it took through the intersection. 
    \item \textbf{CCFtap events}: built on top of the intersection traversal data, a CCFtap event is defined as a \texttt{straight}-labelled traversal paired with a \texttt{left}- or \texttt{right}-labelled traversal in the same intersection over an overlapping time window, subject to a geometric opposite-approach constraint that requires the two vehicles to have entered the intersection from approximately opposite legs. 
\end{itemize}

\subsubsection{Question-Answer Pair Generation}

We generate questions for three reasoning categories:
\begin{itemize}
    \item \textbf{Causality}: questions that ask why a conflict or risk exists.
    \item \textbf{Counterfactual}: questions that ask what may happen under a changed condition.
    \item \textbf{Intent prediction}: questions that ask about the likely future behavior of a road user.
\end{itemize}

These categories are chosen because they require more than object recognition.
Models must reason about interactions between the ego vehicle and other road users while locating the view that contains the relevant evidence.
For example, a question about a vehicle approaching from the back-left should be grounded in \texttt{CAM\_BACK\_LEFT}, not a generic front-facing view.
For answer evaluation, each question can be instantiated in either a multiple-choice format or a free-form format.
The multiple-choice version supports deterministic answer scoring, while the free-form version tests whether models can generate natural explanations without being constrained to predefined options.

\subsubsection{Verification}

The view label is proposed automatically and then manually verified.
Annotators inspect all six camera views and check that the question is answerable, the proposed golden view contains the required evidence, and no other view provides equally direct support.
For unsupported examples, annotators verify that none of the six views directly grounds the question.
They also verify that the reference answer is visually consistent.
Ambiguous samples are revised or removed.
This process balances scalability with annotation reliability.

\subsubsection{Annotator Training, Recruitment, and Compensation}
\label{app:annotation-protocol}

\paragraph{Instructions.}
Annotators received written annotation guidelines before verification.
The instructions define the task (golden-view and answer verification), label definitions, adjudication rules, and quality criteria aligned with Section~\ref{app:details} (Verification).
They state that work involves reviewing public autonomous-driving camera footage only, with no collection of annotators' personal data beyond standard employment records, and no physical or psychological risks beyond routine screen-based work.

\begin{table*}[t]
    \centering
    \small
    \resizebox{\textwidth}{!}{
    \begin{tabular}{lrrrrrrrrrrrr}
    \toprule
    \textbf{Model} & \multicolumn{3}{c}{\textbf{Front}} & \multicolumn{3}{c}{\textbf{Side}} & \multicolumn{3}{c}{\textbf{Rear}} & \multicolumn{3}{c}{\textbf{None / insufficient}} \\
    \cmidrule(lr){2-4} \cmidrule(lr){5-7} \cmidrule(lr){8-10} \cmidrule(lr){11-13}
     & \textbf{View} & \textbf{MC} & \textbf{Free} & \textbf{View} & \textbf{MC} & \textbf{Free} & \textbf{View} & \textbf{MC} & \textbf{Free} & \textbf{View} & \textbf{MC} & \textbf{Free} \\
    \midrule
    Claude & 96.59 & 96.59 & 80.68 & 46.15 & 80.77 & 65.38 & 100.00 & 50.00 & 0.00 & 16.67 & 33.33 & 16.67 \\
    Gemini & 87.50 & 95.45 & 78.41 & 57.69 & 84.62 & 73.08 & 100.00 & 50.00 & 50.00 & 0.00 & 33.33 & 16.67 \\
    GPT-5.4 & 88.64 & 95.45 & 86.36 & 53.85 & 73.08 & 76.92 & 100.00 & 50.00 & 50.00 & 0.00 & 33.33 & 0.00 \\
    InternVL3 & 73.86 & 85.23 & 38.64 & 38.46 & 88.46 & 57.69 & 100.00 & 100.00 & 0.00 & 0.00 & 0.00 & 0.00 \\
    Qwen2.5-VL & 0.00 & 80.68 & 32.95 & 46.15 & 73.08 & 61.54 & 0.00 & 100.00 & 50.00 & 16.67 & 66.67 & 0.00 \\
    Qwen3-VL & 79.55 & 73.86 & 44.32 & 30.77 & 50.00 & 53.85 & 100.00 & 0.00 & 0.00 & 0.00 & 50.00 & 16.67 \\
    \bottomrule
    \end{tabular}
    }
    \caption{Per-region analysis across models. Values are accuracies (\%) for view identification (\textbf{View}), oracle multiple-choice answering (\textbf{MC}), and oracle free-form answering judged by an LLM (\textbf{Free}).}
    \label{tab:per_region_analysis}
    \centering
    \small
    \resizebox{\textwidth}{!}{
    \begin{tabular}{lrrrrrrrrr}
    \toprule
    \textbf{Model} & \multicolumn{3}{c}{\textbf{Causality}} & \multicolumn{3}{c}{\textbf{Counterfactual}} & \multicolumn{3}{c}{\textbf{Intent prediction}} \\
    \cmidrule(lr){2-4} \cmidrule(lr){5-7} \cmidrule(lr){8-10}
     & \textbf{View} & \textbf{MC} & \textbf{Free} & \textbf{View} & \textbf{MC} & \textbf{Free} & \textbf{View} & \textbf{MC} & \textbf{Free} \\
    \midrule
    Claude & 88.46 & 92.31 & 73.08 & 88.57 & 94.29 & 77.14 & 65.71 & 80.00 & 68.57 \\
    Gemini & 75.00 & 96.15 & 75.00 & 82.86 & 91.43 & 74.29 & 74.29 & 77.14 & 71.43 \\
    GPT-5.4 & 82.69 & 92.31 & 82.69 & 82.86 & 91.43 & 82.86 & 62.86 & 74.29 & 71.43 \\
    InternVL3 & 55.77 & 82.69 & 28.85 & 74.29 & 91.43 & 42.86 & 62.86 & 71.43 & 54.29 \\
    Qwen2.5-VL & 5.77 & 82.69 & 40.38 & 8.57 & 85.71 & 31.43 & 20.00 & 65.71 & 40.00 \\
    Qwen3-VL & 59.62 & 69.23 & 44.23 & 80.00 & 74.29 & 45.71 & 60.00 & 54.29 & 42.86 \\
    \bottomrule
    \end{tabular}
    }
    \caption{Per-question-type analysis across models. Values are accuracies (\%) for view identification (\textbf{View}), oracle multiple-choice answering (\textbf{MC}), and oracle free-form answering judged by an LLM (\textbf{Free}).}
    \label{tab:per_type_analysis}
\end{table*}

\begin{table}[t]
    \centering
    \small
    \resizebox{\columnwidth}{!}{
    \begin{tabular}{lrrrrr}
    \toprule
    Model & G$\checkmark$\,/\,MC$\checkmark$ & G$\checkmark$\,/\,MC$\times$ & G$\times$\,/\,MC$\checkmark$ & G$\times$\,/\,MC$\times$ & Total \\
    \midrule
    Claude & 97 & 3 & 12 & 10 & 122 \\
    Gemini & 90 & 4 & 19 & 9 & 122 \\
    GPT-5.4 & 89 & 5 & 17 & 11 & 122 \\
    InternVL3 & 67 & 10 & 33 & 12 & 122 \\
    Qwen2.5-VL & 10 & 3 & 86 & 23 & 122 \\
    Qwen3-VL & 56 & 24 & 25 & 17 & 122 \\
    All models & 409 & 49 & 192 & 82 & 732 \\
    \bottomrule
    \end{tabular}
    }
    \caption{Outcome counts on the 122-question benchmark, decomposed by golden-view selection (G) and oracle multiple-choice accuracy (MC). Each row sums to 122.}
    \label{tab:golden_mc_counts}
\end{table}

\begin{table}[t]
    \centering
    \small
    \resizebox{\columnwidth}{!}{
    \begin{tabular}{lrrrrr}
    \toprule
    Model & G$\checkmark$\,/\,Free$\checkmark$ & G$\checkmark$\,/\,Free$\times$ & G$\times$\,/\,Free$\checkmark$ & G$\times$\,/\,Free$\times$ & Total \\
    \midrule
    Claude & 81 & 19 & 8 & 14 & 122 \\
    Gemini & 75 & 19 & 15 & 13 & 122 \\
    GPT-5.4 & 79 & 15 & 18 & 10 & 122 \\
    InternVL3 & 29 & 48 & 20 & 25 & 122 \\
    Qwen2.5-VL & 7 & 6 & 39 & 70 & 122 \\
    Qwen3-VL & 38 & 42 & 16 & 26 & 122 \\
    All models & 309 & 149 & 116 & 158 & 732 \\
    \bottomrule
    \end{tabular}
    }
    \caption{Outcome counts on the 122-question benchmark, decomposed by golden-view selection (G) and oracle free-form accuracy judged by an LLM (Free). Each row sums to 122.}
    \label{tab:golden_free_counts}
\end{table}

\paragraph{Recruitment and payment.}
We recruited graduate-student annotators through standard recruitment channels. Compensation was provided as hourly pay at rates consistent with comparable graduate research-assistant annotation work.

Annotators completed a short training session on the guidelines and pilot examples before independent labeling.

\paragraph{Data consent.}
All scene images come from the public nuScenes dataset~\citep{caesar_NuScenesMultimodalDataset_2020}, used under its published terms of use; we do not collect new in-the-wild video from annotators or bystanders.
Annotators were informed that their labels would be used to build and release a research benchmark derived from nuScenes, and that they should not share raw annotation materials outside the project.

\subsection{Qualitative Case Studies}
\label{app:qualitative-cases}

We summarize qualitative failure modes through aggregate case-study tables that decompose each prediction by whether the model selects the correct view and whether it answers correctly.

\paragraph{Analysis.}
\Cref{tab:golden_mc_counts,tab:golden_free_counts} show that answer correctness and source identification are only partially aligned.
Across all models, 192 out of 732 multiple-choice trials fall into the wrong-view/correct-answer quadrant, meaning that more than one quarter of model-example pairs produce the right option despite selecting the wrong evidence source.
This pattern is especially pronounced for Qwen2.5-VL, where 86 multiple-choice answers are correct despite an incorrect view, suggesting that answer-only evaluation would substantially overestimate grounded reasoning for this model.
Free-form answering reduces but does not remove this effect: 116 trials remain wrong-view/free-form-correct, while 149 trials are correct-view/free-form-wrong, indicating that free generation exposes both evidence misattribution and downstream reasoning failures.

\Cref{tab:per_region_analysis} further shows that view localization is not uniformly difficult across camera regions.
Most models perform much better on front and rear views than on side views, where adjacent-camera ambiguity and partial actor visibility are more common.
The \textsc{None}/insufficient-evidence cases are particularly challenging: even strong proprietary models rarely abstain correctly, which supports including unsupported examples as a diagnostic stress test rather than treating all questions as answerable by construction.
\Cref{tab:per_type_analysis} shows a complementary pattern across reasoning types.
Oracle multiple-choice accuracy remains relatively high for many models, but view identification drops on intent-prediction questions for several systems, suggesting that forecasting-oriented questions require models to identify subtle interaction evidence rather than only salient objects.
Together, these tables support the central claim of the benchmark: final-answer accuracy alone conflates visual-source localization, answer reasoning, and abstention behavior.

\section{Prompts}
\label{app:prompts}

We provide the prompt templates used for model inference and for grading free-form answers.
Each instance is constructed from a benchmark item: the natural-language question $q$, metadata such as task type $\tau$ and horizon $h$, and---depending on the setting---either a single reference image or the set of synchronized multi-view frames at one nuScenes keyframe.
Bracketed placeholders in the boxes below (e.g., \texttt{<question>}) denote fields filled per instance at runtime.
The candidate camera set is
\begin{align*}
    \mathcal{C} = \{ 
        &\;\textsc{CamFront},\;\;\;\;\;\;\textsc{CamFrontLeft},\\
        &\;\textsc{CamFrontRight},\;\;\;\;\textsc{CamBack},\\
        &\;\textsc{CamBackLeft},\;\;\;\;\textsc{CamBackRight},\\
        &\;\textsc{None} 
    \}
\end{align*}
where \textsc{None} indicates that no provided view adequately supports the question.
In the multiple-choice setting, the model must select one option identifier from a question-specific list $\mathcal{O}$; we score predictions by exact match to the gold identifier.
Free-form responses are evaluated with a separate LLM judge (Section~\ref{app:prompts:judge}).

\paragraph{Multimodal message format.}
For API-based models, the user turn contains the textual prompt followed by each candidate view preceded by its channel name.
For local vision--language models that accept an ordered image list, the user prompt states the channel order explicitly; images are inserted in that order (excluding \textsc{None}).
All inference prompts request structured JSON; schemas appear in Section~\ref{app:prompts:schemas}.
We use temperature $0$ for view-selection and joint-prediction calls unless noted otherwise in the main text.

\subsection{Camera View Selection}
\label{app:prompts:golden-view}

The model observes all synchronized views at timestamp $t$ and predicts the channel $c^\star \in \mathcal{C}$ whose field of view best exhibits the evidence required to answer $q$.

\begin{TemplateBox}{View selection --- system}
You are selecting the single best camera view from synchronized nuScenes images.\\[0.5em]
Rules:\\
- Use only the provided images and question text.\\
- Choose exactly one best\_camera\_channel from the provided candidate channels.\\
- Prefer the single camera that most directly shows the visual evidence needed to answer the benchmark question.\\
- If none of the provided camera views directly grounds the question, choose NONE\_OF\_THE\_ABOVE.\\
- If several views are useful, choose the one with the clearest visibility of the key object, action, or constraint.\\
- Keep the rationale short and grounded in visible evidence.
\end{TemplateBox}

\begin{TemplateBox}{View selection --- user}
Question ID: <question\_id>\\
Task type: <task\_type>\\
Original question text: <original\_question>\\
Benchmark question: <question>\\
Annotated frame: <frame\_index>\\[0.5em]
Choose the single camera channel that best shows the scenario or visual evidence needed to answer the benchmark question from this synchronized timestamp. Choose NONE\_OF\_THE\_ABOVE only if none of the provided camera views directly grounds the question.\\[0.5em]
Candidate camera channels:\\
- <channel\_1>\\
- <channel\_2>\\
\ldots\\[0.5em]
Return the best\_camera\_channel, a short rationale, visible evidence, and a confidence score.
\end{TemplateBox}

\subsection{Grounded QA with Oracle View}
\label{app:prompts:oracle}

The model receives only the human-annotated gold-view image $I_{c_{\text{gold}}}$ and answers $q$.
This setting isolates reasoning given correct visual grounding.

\subsubsection{Multiple-choice}
\label{app:prompts:oracle-mc}

\begin{TemplateBox}{Oracle QA (MC) --- system}
You are answering a single-frame driving benchmark question.\\[0.5em]
Rules:\\
- Use only the provided image and question text.\\
- Do not use scene-level future knowledge beyond what can reasonably be inferred from the current frame.\\
- Choose exactly one answer\_id from the provided options.\\
- If the image does not support a confident decision, choose insufficient\_evidence.\\
- Keep the rationale short and grounded in visible evidence.
\end{TemplateBox}

\begin{TemplateBox}{Oracle QA (MC) --- user}
Question ID: <question\_id>\\
Task type: <task\_type>\\
Time horizon: <horizon> seconds\\
Question: <question>\\[0.5em]
Options:\\
- <option\_id\_1>: <option\_text\_1>\\
- <option\_id\_2>: <option\_text\_2>\\
\ldots\\[0.5em]
Return the best answer\_id, a short rationale, visible evidence, and a confidence score.
\end{TemplateBox}

\subsubsection{Free-form}
\label{app:prompts:oracle-freeform}

\begin{TemplateBox}{Oracle QA (free-form) --- system}
You are answering a single-frame driving benchmark question.\\[0.5em]
Rules:\\
- Use only the provided image and question text.\\
- Do not use scene-level future knowledge beyond what can reasonably be inferred from the current frame.\\
- Write a concise free-form answer in your own words. Do not pick from a fixed option list.\\
- If the image does not support a confident decision, say clearly that there is insufficient evidence.\\
- Keep the rationale short and grounded in visible evidence.
\end{TemplateBox}

\begin{TemplateBox}{Oracle QA (free-form) --- user}
Question ID: <question\_id>\\
Task type: <task\_type>\\
Time horizon: <horizon> seconds\\
Question: <question>\\[0.5em]
No multiple-choice options are provided. Return a concise free-form answer, a short rationale, visible evidence, and a confidence score.
\end{TemplateBox}

\subsection{Joint View Selection and Answering}
\label{app:prompts:combined}

In a single forward pass the model predicts $(c^\star, a)$: the supporting view and the answer to $q$, along with a brief rationale tied to $c^\star$.
The multimodal input format follows Section~\ref{app:prompts:golden-view}.

\subsubsection{Multiple-choice}
\label{app:prompts:combined-mc}

\begin{TemplateBox}{Joint prediction (MC) --- system}
You are answering a synchronized multi-camera nuScenes driving benchmark question.\\[0.5em]
Rules:\\
- Use only the provided images and question text.\\
- Do not use scene-level future knowledge beyond what can reasonably be inferred from the current timestamp.\\
- Return JSON only with keys Golden\_view, Answer, and Rationale.\\
- Golden\_view: choose exactly one camera channel from the provided candidates that best shows the visual evidence needed to answer the question. Use NONE\_OF\_THE\_ABOVE only if none of the provided views directly grounds the question.\\
- Answer: choose exactly one answer\_id from the provided options. Use insufficient\_evidence if no view supports a confident decision.\\
- Rationale: briefly explain which visible evidence in your chosen Golden\_view supports your Answer.
\end{TemplateBox}

\begin{TemplateBox}{Joint prediction (MC) --- user}
Question ID: <question\_id>\\
Task type: <task\_type>\\
Original question text: <original\_question>\\
Benchmark question: <question>\\
Time horizon: <horizon> seconds\\
Annotated frame: <frame\_index>\\[0.5em]
From the synchronized camera images at this timestamp:\\
1. Pick Golden\_view: the single camera channel that best shows the visual evidence for answering the question (or NONE\_OF\_THE\_ABOVE if none apply).\\
2. Pick Answer: the best answer\_id from the options below.\\
3. Write Rationale: short explanation grounded in what is visible in Golden\_view.\\[0.5em]
Candidate camera channels:\\
- <channel\_1>\\
\ldots\\[0.5em]
Options:\\
- <option\_id\_1>: <option\_text\_1>\\
\ldots\\[0.5em]
Return JSON with exactly these keys: Golden\_view, Answer, Rationale.
\end{TemplateBox}

\subsubsection{Free-form}
\label{app:prompts:combined-freeform}

\begin{TemplateBox}{Joint prediction (free-form) --- system}
You are answering a synchronized multi-camera nuScenes driving benchmark question.\\[0.5em]
Rules:\\
- Use only the provided images and question text.\\
- Do not use scene-level future knowledge beyond what can reasonably be inferred from the current timestamp.\\
- Return JSON only with keys Golden\_view, Answer, and Rationale.\\
- Golden\_view: choose exactly one camera channel from the provided candidates that best shows the visual evidence needed to answer the question. Use NONE\_OF\_THE\_ABOVE only if none of the provided views directly grounds the question.\\
- Answer: write a concise free-form answer in your own words. Do not pick from a fixed option list. If no view supports a confident decision, say clearly that there is insufficient evidence.\\
- Rationale: briefly explain which visible evidence in your chosen Golden\_view supports your Answer.
\end{TemplateBox}

\begin{TemplateBox}{Joint prediction (free-form) --- user}
Question ID: <question\_id>\\
Task type: <task\_type>\\
Original question text: <original\_question>\\
Benchmark question: <question>\\
Time horizon: <horizon> seconds\\
Annotated frame: <frame\_index>\\[0.5em]
From the synchronized camera images at this timestamp:\\
1. Pick Golden\_view: the single camera channel that best shows the visual evidence for answering the question (or NONE\_OF\_THE\_ABOVE if none apply).\\
2. Write Answer: a concise free-form answer in your own words (no option list).\\
3. Write Rationale: short explanation grounded in what is visible in Golden\_view.\\[0.5em]
Candidate camera channels:\\
- <channel\_1>\\
\ldots\\[0.5em]
Return JSON with exactly these keys: Golden\_view, Answer, Rationale.
\end{TemplateBox}

\subsection{LLM Judge for Free-Form Answers}
\label{app:prompts:judge}

Free-form predictions are graded by a separate vision--language model prompted as a semantic evaluator.
The judge receives $q$, the gold reference answer, the model's free-form answer, and optionally the model's rationale.
When a frame is provided, we use the image at the predicted view for joint-setting outputs and the gold-view image for oracle-setting outputs, so that grading reflects the visual context the answer claims to rely on.
Multiple-choice predictions are \emph{not} sent to this judge; they are scored by identifier-level exact match.

\begin{TemplateBox}{Answer grading --- system}
You are grading a free-form driving benchmark answer. Decide whether the predicted answer is correct with respect to the gold reference answer for the same question.\\[0.5em]
Evaluation rules:\\
1. Primary criterion: semantic equivalence to the gold reference answer.\\
2. Ignore minor wording differences, paraphrases, singular/plural variation, and equivalent expressions.\\
3. Be strict on factual content; be lenient on phrasing.\\
4. When an image is provided, the predicted answer must also be plausibly supported by visible evidence in that frame. If the text matches gold but is clearly unsupported by the image, mark incorrect.\\
5. If both predicted and gold answers clearly indicate insufficient evidence, and the image does not contradict that, mark correct (verdict: insufficient\_evidence\_ok).\\
6. If the predicted answer contradicts the gold answer, mark incorrect.\\[0.5em]
Return JSON only with:\\
- correct: 1 if correct, 0 if incorrect\\
- verdict: one of correct, incorrect, insufficient\_evidence\_ok\\
- reason: short explanation grounded in the comparison (and image when provided)
\end{TemplateBox}

\begin{TemplateBox}{Answer grading --- user}
Question ID: <question\_id>\\
Task type: <task\_type>\\
Question: <question>\\[0.5em]
Gold reference answer: <gold\_answer>\\
Predicted answer: <predicted\_answer>\\[0.5em]
Predicted rationale: <predicted\_rationale>\\[0.5em]
Return JSON with keys: correct (0 or 1), verdict, and reason.
\end{TemplateBox}

\subsection{Structured Response Formats}
\label{app:prompts:schemas}

Models are instructed to return JSON objects with the fields below.
Enumerated fields are instantiated per question (e.g., option identifiers or channels in $\mathcal{C}$).
Providers that support strict JSON schema use these definitions directly; otherwise the required keys are repeated in the system prompt.

\begin{TemplateBox}{Schema: view selection}
\{\\
\ \ "best\_camera\_channel": "<enum over $\mathcal{C}$>",\\
\ \ "confidence": <float $\in [0,1]$>,\\
\ \ "visible\_evidence": ["<string>", \ldots],\\
\ \ "rationale": "<string>"\\
\}
\end{TemplateBox}

\begin{TemplateBox}{Schema: oracle QA (multiple-choice)}
\{\\
\ \ "answer\_id": "<enum over $\mathcal{O}$>",\\
\ \ "confidence": <float $\in [0,1]$>,\\
\ \ "visible\_evidence": ["<string>", \ldots],\\
\ \ "rationale": "<string>"\\
\}
\end{TemplateBox}

\begin{TemplateBox}{Schema: oracle QA (free-form)}
\{\\
\ \ "answer": "<string>",\\
\ \ "confidence": <float $\in [0,1]$>,\\
\ \ "visible\_evidence": ["<string>", \ldots],\\
\ \ "rationale": "<string>"\\
\}
\end{TemplateBox}

\begin{TemplateBox}{Schema: joint prediction (multiple-choice)}
\{\\
\ \ "Golden\_view": "<enum over $\mathcal{C}$>",\\
\ \ "Answer": "<enum over $\mathcal{O}$>",\\
\ \ "Rationale": "<string>"\\
\}
\end{TemplateBox}

\begin{TemplateBox}{Schema: joint prediction (free-form)}
\{\\
\ \ "Golden\_view": "<enum over $\mathcal{C}$>",\\
\ \ "Answer": "<string>",\\
\ \ "Rationale": "<string>"\\
\}
\end{TemplateBox}

\begin{TemplateBox}{Schema: answer grading}
\{\\
\ \ "correct": 0 \textbar\ 1,\\
\ \ "verdict": "correct" \textbar\ "incorrect" \textbar\ "insufficient\_evidence\_ok",\\
\ \ "reason": "<string>"\\
\}
\end{TemplateBox}

\section{AI Disclosure}
\label{app:ai-disclosure}

AI writing assistance was used only for proofreading and formatting support. All scientific content, experiments, analyses, and conclusions were produced and verified by the authors.

\end{document}